\documentclass[a4paper,11pt]{article}
\usepackage{xcolor}
\usepackage{hyperref}
\hypersetup{colorlinks=true,linkcolor=blue,urlcolor=blue}
\usepackage{graphicx}
\usepackage{epstopdf}
\usepackage{epigraph} 
\begin{document}  
\title{Linear algorithm for solution $n$-Queens Completion problem}
\author{E. Grigoryan, \\Independent researcher, Marseille, France \\ericgrig@gmail.com} 
\maketitle
\setlength{\epigraphwidth}{0.5\textwidth}  
\epigraph{Publication is dedicated to Yuri Milner,}{\textit{wonderful person - the new Nobel of our time \ldots}}
\begin{abstract}
1. A linear algorithm is described for solving the $n$-Queens Completion problem for an arbitrary composition of $k$ queens, consistently distributed on a chessboard of size $n \times n$. Two important rules are used in the algorithm: a) the rule of sequential risk elimination for the entire system as a whole; b) the rule of formation of minimal damage in the given selection conditions. For any composition of $k$ queens $(1 \leq k < n$), a solution is provided, or a decision is made that this composition can't be completed. The probability of an error in making such a decision does not exceed 0.0001, and its value decreases, with increasing $n$.  2. It is established that the average time, required for the queen to be placed on one row, decreases with increasing value of $n$. 3. A description is given of two random selection models and the results of their comparative analysis. 4. A model for organizing the Back Tracking procedure is proposed based on the separation of the solution matrix into two basic levels. Regression formulas are given for the dependence of basic levels on the value of $n$.  5. It was found that for  $n=( 7,...,100000 )$  the number of solutions in which the Back Tracking procedure has never been used exceeds 35\%. Moreover, for $n =( 320,...,22500 )$, the number of such cases exceeds 50\% .  6. A quick algorithm for verifying the correctness of $n$-Queens problem solution or arbitrary composition of $k$ queens is given.
\end{abstract}
\section{Introduction}
\label{Introduction}
Among the various formulations of the $n$-Queens problem, the $n$-Queens Completion problem has a special position due to its complexity. In the work, Gent at all \cite{Gent-2017} showed that $n$-Queens Completion is both $NP$-Complete and $\sharp P$-Complete. It is assumed that the solution of this problem may open the way to solve some other problems from the set of $NP$-Complete.\
\newline\indent The problem is formulated as follows. There is a composition of $k$ queens, which are consistency distributed on a chessboard of size $n \times n$. It is required to prove that this composition can be completed to a complete solution, and give at least one solution, or to prove that such a solution does not exist. Here, by consistency, we mean a composition of k queens for which three conditions of the problem are fulfilled: in each row, each column, and also on the left and right diagonals passing through the cell where the queen is located, no more than one queen is located. The problem in this form was first formulated by Nauk \cite{Nauk-1850} in 1850.
\subsection{Definitions}
	Hereinafter, we will denote the size of the side of the chessboard by the symbol $n$. A solution will be called complete if all $n$ queens are consistently placed on a chessboard. All other solutions, when the number $k$ of correctly placed queens is less than $n$ - we will call the composition. We call a composition of $k$ queens positive if it can be completed. Accordingly, a composition that can't be completed is called negative. As an analogue of a "chessboard"\ of size $n \times n$, we will also consider a "solution matrix"\ of size $n \times n$. Research based on Matlab scripting language.\
\newline\indent The study was based on computational simulation. To test different hypothesis, we carried out computational experiments in a wide range of values $n$= (10, 20, 30, 40, 50, 60, 70,  80, 90, 100, 200, 300, 500,  800, 1000, 3000, 5000, 10000, 30000, 50000, 80000, $10^5$,  $3\ast10^5$,  $5\ast10^5$,  $10^6$,   $3\ast10^6$,  $5\ast10^6$, $10^7$,  $3\ast10^7$,  $5\ast10^7$,  $8\ast10^7$,  $10^8)$ and generated, depending on the value of $n$, large enough samples for analysis. We call such a list a "basic list of $n$ values"\ for computational experiments. All calculations were carried out on a regular computer. At the time of assembly (early 2013), it was a fairly successful configuration: CPU - Intel Core $i7$-3820, 3.60 GH, RAM-32.0 GB, GPU- NVIDIA Ge Forse GTX 550 Ti, Disk device- ATA Intel SSD, SCSI, OS- 64-bit Operating system Windows 7 Professional. We call this configuration simply - desktop-13.
\section{Data preparation}
The algorithm begins by reading a file that contains a one-dimensional array of data on the distribution of an arbitrary composition of $k$ queens. It is assumed that the data is prepared in the following way. Let there be a zeroed array $Q (i)$ = 0, $i$ = (1,...,$n$), where the indices of the cells of this array correspond to the row indices of the solution matrix. If in some arbitrary row $i$ of the solution matrix there is a queen in position $j$, then the assignment $Q(i) = j$ is performed. Thus, the composition size $k$, will be equal to the number of non zero cells of the array $Q$. (For example, $Q$ = (0, 0, 5, 0, 4, 0, 0, 3, 0, 0) means that we consider a composition of $k$ = 3 queens on matrix $n$ = 10, where the queens are located in the 3rd, 5th and 8th rows, respectively, at the positions: 5, 4, 3).
\section{Algorithm for validation the n-Queens Problem solution}
For research, we need an algorithm that would allow us to determine the correctness of the solution of the $n$-Queens problem in a short time. Controlling the location of queens in each row and each column is simple. The question is about diagonal constraints. We could build an effective algorithm for such an accounting if we could associate each cell of the solution matrix with a certain cell of a certain control vector that would uniquely characterize the influence of diagonal constraints on the cell. Then, based on whether the cell of the control vector is free or closed, we can judge whether the corresponding cell of the solution matrix is free or excluded. Such an idea was first used by Sosic and Gu \cite{Sosic-1990} to take into account and accumulate the number of conflict situations between different positions of queens. We use a similar idea in the algorithm presented below, but only to take into account whether the cell of the solution matrix is free or excluded. In Figure 1, as an example, shows a 8 x 8 chessboard above which a sequence of 24 cells is located. Consider the first 15 cells as elements of the control vector $D1$. The projections of all left diagonals from any cell of the solution matrix fall into one of the cells of the control vector $D1$. In fact, all such projections are located inside two parallel line segments, one of which connects the matrix cell $(8,1)$ to the first cell of the vector $D1$, and the second connects the matrix cell $(1, 8)$ to the 15th cell of the control vector $D1$. We give a similar definition for right diagonal projections. For this we move right the reference point from cell 1 to cell 9, and consider a sequence of 16 cells as elements of the control vector $D2$ (in the figure, these are cells from the 9th to the 24th). The projections of all right diagonals from any cell of the solution matrix will fall into one of the cells of this control vector, starting from the 2nd cell through the 16th (in the figure, from 10th to 24th). Here, all such projections are located between two segments of parallel lines — the segment connecting the cell (8, 8) of the solution matrix with cell 16 of the vector $D2$ (cell 24 in the figure) and the segment connecting the cell (1, 1) of the solution matrix with the cell 2 of the control vector $D2$ (cell 10 in the figure). The projections of all cells of the solution matrix lying on the same left diagonal fall into the same cell of the left control vector $D1$, respectively, the projections of all cells of the solution matrix lying on the same right diagonal fall into the same cell of the right control vector $D2$. Thus, these two control vectors $D1$ and $D2$, allow full control of all diagonal inhibitions for any cell of the solution matrix.	
\begin{figure}[ht]
\center\includegraphics[width=0.9\textwidth]{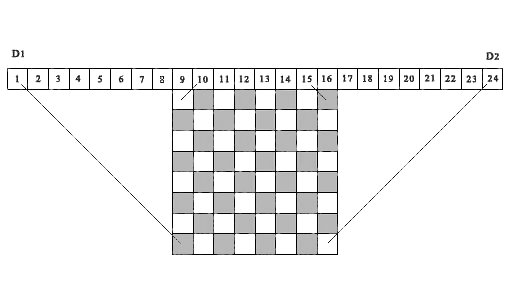} 
\caption{Demo example of the diagonal projections of the matrix cells to the corresponding cells of the control arrays $D1$ and $D2$. ($n$ = 8)}
\label{Fig-1}
\end{figure}
\\
\newline\indent It is important to note that the idea of using diagonal projections onto cells of control vectors to determine whether a cell of a solution matrix with coordinates $(i,j)$ is free or busy was also later implemented by Richards \cite{Richards-1997}. This publication provides one of the fastest recursive search algorithms for all solutions, based on operations with a bit mask. An important difference is that this algorithm is designed for the sequential search of all solutions, starting from the first row of the solution matrix - down, or from the last row of the matrix - up. The algorithm we proposed is based on the condition that the choice of the number of each row for the location of the queen must be arbitrary. For the considered algorithm, this is of fundamental importance. Note that the above figure 1, we built by analogy with what is published in this paper.
\\
\newline A program to check whether a given solution of the $n$-Queens problem is correct, or whether a given composition of $k$ queens is true, is as follows.
\\
\newline\indent 1. To control diagonal constraints, we must create two arrays $D1(1:n2)$ and $D2(1:n2)$, where $n2 = 2\ast n$, and array $B(1: n$) to control the occupancy of the columns of the solution matrix. Set to zero these three arrays.
\\
\newline\indent 2. We introduce the counter of the number of correctly installed queens $(totPos$ = 0). Consistently, in a cycle, starting from the first row, we consider all the queens positions provided. If $Q(i)>$ 0, then based on the index of row $i$ and the index of the position of the queen in this row $j = Q(i)$, we form the corresponding indices for the control arrays $D1(r)$ and $D2(t)$:
\\
\newline\indent\indent $r$ = $n$ + $j$ – $i$
\\
\newline\indent\indent $t$ = $j + i$
\\
\newline\indent 3. If all conditions $D1(r)$ = 0, $D2(t)$ = 0, $B(j)$ = 0 are satisfied, this will mean that the cell $(i,j)$ is free and does not fall into projection zone of diagonal restrictions formed by previously established queens. Placing the queen in this position is correct. If at least one of these conditions is not fulfilled, then the choice of such a position will be erroneous, and the solution will be erroneous.
\\
\newline\indent 4. If the solution is correct, then we increment the counter of the number of correctly installed queens and close the corresponding cells of the control arrays: ($totPos$ = $totPos$ + 1, $D1(r)$ = 1, $D2(t)$ = 1, $B(j)$ = 1). Thus, we close all cells of the column $(j)$ and those cells of the solution matrix that are located along the left and right diagonals intersecting in the cell $(i,j)$.
\\
\newline\indent 5. Repeat the verification procedure for all remaining positions.
\\
\newline\indent Perhaps, this is one of the fastest algorithm for validation the $n$-Queens problem solution. The verification time of ${10}^6$ positions on the desktop-13 is 0.175 seconds. Which corresponds to the time to press the Enter key.
\section{Algorithm description}
The $n$-Queens Completion problem is a classic non-deterministic problem. The main difficulty of its solution is connected with the question of selecting the row index and position index in this row, in conditions when the state space is huge. When searching for all possible solutions, such a problem does not arise, we must consider all valid search branches in the state space, and the order in which they are considered does not matter. However, when an arbitrary composition of $k$ queens needs to be completed, then in this case we need an algorithm for selecting row and column indices that adequately perceives the existing composition and leads to a solution faster than others. In this project, we decided the question of choice on the basis of the following general position - if we can't formulate conditions that give preference to any row or any position in this row over others, then we use a random selection algorithm based on evenly distributed random numbers. A similar random selection method for solving problems in which the state space is huge is quite natural. One of the editions of the Proceedings of a DIMACS Workshop series \cite{DIMACS-1997} was completely devoted to the use of random selection in developing algorithms for solving complex problems. The correct implementation of the random selection algorithm can be a fairly productive approach, although this is a necessary but not sufficient condition for the completion of the solution. The article Sosic and Gu \cite{Sosic-1990} is one of the first studies to use a random selection algorithm to solve the $n$-Queens Problem. The algorithm they examined is based on a fairly simple and concise idea. Let there be a sequence of numbers from 1 to $n$, which are randomly rearranged. Such a set of numbers has an important property. It consists in the fact that no matter how these numbers are distributed on different rows of the solution matrix as the positions of the queen, the first two rules in the statement of the problem will always be fulfilled: no more than one queen will be located in each row and each column. However, only a part of the positions thus obtained will be free from diagonal restrictions. The other part will be in a state of "conflict" with previously established queens. To overcome this situation, the authors used the method of comparing and interchanging conflicting positions in order to obtain a complete solution. In our proposed algorithm, two random selection methods are used at various stages of the solution. Moreover, at the final stage, the decision is formed mainly on the basis of established rules where the share of random selection is negligible. Conflict situations are impossible in this algorithm, since at each step of solution, the queen is installed in the cell of the row if the cell is free from restriction.
\subsection{Model for organizing the Back Tracking procedure}
In a solution process may arise a situation when a sequential chain of solutions leads to a dead end. This is a "genetic" property of non-deterministic problems. In this case, we need to go back to one of the previous steps, restore the state of the task in accordance with this level and start again solution process from this position. The question is which of the previous levels should be returned and how many such levels should be (by level, we mean a certain step in solving the problem with a given number of correctly installed queens). Obviously, choosing a solution level to go back is just as relevant as choosing a row index or a position index in that row. Therefore, regardless of the approach to solving this problem, it is necessary to first determine the number of basic levels for returning back, as well as the mechanism and conditions for returning to one of these levels. In our proposed algorithm, we divide the solution matrix into three basic levels. These are the return points. If, as a result of the solution, a deadlock occurs, then, depending on the parameters of the task, we return back to one of these three basic levels. The first base level ($baseLevel_1)$ corresponds to the state when the data verification of the composition is completed. This is the beginning of the program. The values of the following two base levels ($baseLevel_2$ and $baseLevel_3$) depend on the size of the matrix $n$. The empirical dependence of these basic values on the size of the solution matrix was established on the basis of a large number of computational experiments. For a more accurate representation of this dependence, we divided the entire considered interval from 7 to $10^8$ into two parts. Let $u$ = $lg(n)$, then
\\
if $n < 30000$
\\
\newline\indent $baseLevel_2$ = $n - round(12.749568*u3 - 46.535838*u2 + 120.011829*u - 89.600272)$
\\ 
\newline\indent $baseLevel_3$ = $n - round( 9.717958*u3 – 46.144187*u2 + 101.296409*u – 50.669273)$
\\
else
\\
\newline\indent $baseLevel_2$ = $n - round(-0.886344*u3  + 56.136743*u2 +  146.486415*u +  227.967782)$
\\
\newline\indent $baseLevel_3$ = $n - round(14.959815*u3 – 253.661725*u2 + 1584.713376*u – 3060.691342)$
\\
\subsection{Block structure}
The algorithm is built in the form of a sequence of five event blocks, where each event is associated with the implementation of a certain part of problem solution. The algorithms in each block are different from each other. Only three blocks serve to form the solution, the remaining two blocks are preparatory. The choice of the block number from which the calculations begin depends on the value of $n$ and on the results of comparing the composition size $k$ with the values of $baseLeve_2$ and $baseLevel_3$. An exception is the interval of values $n$ = (7,...,99), which can be called a "turbulent zone" due to the peculiarities of the algorithm behavior in this section. For values $n$ = (7,...,49), regardless of the composition size, after entering and controling data, the calculations begin from the 4th block. For values $n$ = (50,...,99), depending on the composition size, calculations begin either from the second block or from the fourth. At each stage of solving the problem, only those positions in the row are considered that don't fall into the zone of restrictions created by previously established queens. It is these positions that are called free.
Let us briefly describe what calculations are performed in each of these five blocks.
\subsection{Begining of the algorithm}
Data is entered and the composition is checked for correctness. At each verification step, the cells of the control arrays are changed. A count of the number of correctly installed queens is kept. If there are no errors in the composition, the solution continues, otherwise an error message is displayed. After verification is complete, copies of the main arrays are created for their reuse at this level. After that, control is transferred to $Block$-1.
\subsection{Block-1}
The beginning of the formation of the search branch. We consider $k$ queens located on a chessboard as a starting position. It is required to continue to complete this composition and place the queens on a chessboard until their total number is equal to $baseLevel_2$. The algorithm used here is called $randSet$ \& $randSet$. This is due to the fact that here we are constantly comparing two random lists of indices, in search of pairs free from the corresponding diagonal restrictions. To do this, the following actions are performed:
\\
\newline\indent a) two lists are formed: a list of free row indices and a list of free column indices;
\\
\newline\indent b) perform random permutation of numbers in each of these lists;
\\
\newline\indent c) in a loop, each consecutive pair of values $(i, j)$, where index $(i)$ is selected from the list of free row indices and index $(j)$ from the list of free column indices, is considered as a potential queen position, and it is checked whether this position in the projection area of diagonal exceptions.
\\
\newline If the rule of diagonal exceptions is not violated, then the position is considered correct, and the queen is placed in this position. After that, the counter for the number of correctly installed queens is incremented, and the corresponding cells of the control arrays are changed. If the position $(i, j)$ falls into the zone of diagonal restrictions formed by the queens established earlier, then nothing changes and the transition to consideration of the next pair of values takes place.
When the comparison cycle of all pairs is completed, based on the remaining indices, a list of indices of the remaining free rows and free columns is again formed, and this procedure is repeated until the total number of correctly placed queens ($totPos$) is equal to or exceeds the limit value of  $baseLevel_2$ . Once this condition is met, control is transferred to $Block$-2. If it turns out that as a result of the search, a situation arose that from the entire list of indices of the remaining free rows and free columns, none of the pairs is suitable for the location of the queen, then in this case, the original values of the control arrays are restored based on previously generated copies , and control is transferred to the beginning of $Block$-1 for re-counting.
\subsection{Block-2}
This block serves as a preparatory stage for the transition to $Block$-3. At this level, the number of remaining free rows ($freeRows$) is significantly less than $n$. This allows us to transfer events from the original $n \times n$ matrix to a smaller matrix $L$(1: $freeRows$, 1: $freeRows$). Moreover, based on the information about the remaining free rows and free columns in the original solution matrix, zeros are written to the corresponding cells of the array $L$, indicating that these cells are free. With such a "projection" transition, the correspondence of the row and column indices of the new matrix with the corresponding indices of the original matrix is preserved. It is important to note that although, in the process of solving this problem, all events unfold on the initial matrix of size $n \times n$, and such a matrix is the main arena of action, in reality such a matrix is not created,  and we only create control arrays of accounting for row indices $A$ (1: $n$) and column indices $B$ (1: $n$) of this matrix.
Along with the $L$ array, two working arrays $rAr$ and $tAr$ are also formed in this block to save the corresponding indices of the control arrays $D1$ and $D2$. This is due to the fact that when we install the next queen in the cell ($i$, $j$) of the initial matrix of size $n \times n$, then after that we must exclude the cells of the array $L$ that fall into the projection zone of the diagonal exceptions of the original "large" array. Since the control of diagonal constraints is carried out only within the initial matrix of size $n \times n$, the presence of working arrays $rAr$ and $tAr$ allows us to maintain correspondence and translate forbidden cells to the limits of array $L$. This greatly simplifies the accounting of excluded positions.
After the preparatory work is completed in this block, copies of the main arrays are created for reuse at this level, and control is transferred to $Block$-3.
\subsection{Block-3}
In this block, the formation of search branch continues on the basis of the data prepared in the previous block. The number of rows in which queens are correctly set is equal to or greater than $baseLevel_2$. We need to continue solution until the number of queens installed is equal to $baseLevel_3$. The $rand$ \& $rand$ solution search algorithm is used here, i.e. to form a queen's position, instead of a list of free indexes, only two indexes are used, a random index value of a free row and a random index value of a free position in this row. This procedure is repeated cyclically until the total number of queens placed is equal to the value of $baseLevel_3$. Once this condition is met, control is transferred to $Block$-4. If, as a result of the calculations, the search branch is deadlocked, then this search branch is closed and a return is made to the beginning of event 3, from where the calculations are repeated again. For this, the initial values of all control arrays are restored.
\subsection{Block-4}
In this block, data is prepared for the transfer of control to $Block$-5. To this step, after completing the procedure in $Block$-3, the number of free rows ($nRow$) has become even less. Therefore, it is also beneficial to transfer events from a larger array to a smaller array. This approach gives us the opportunity to quickly determine the necessary characteristics for the remaining rows that we need at this stage. Of particular importance is the fact that on the basis of such an array, it is possible to predict the prospectivity of the search branch for many steps forward without having to complete the calculations. The condition is quite simple. If it turns out that among the remaining free rows there is a row in which there is no free position, then the search branch under consideration is closed and control is transferred to one of the lower-level blocks. The preparatory actions carried out here are in many respects similar to what was done in $Block$-2. Based on the original indices of free rows and free columns, a new 2-dimensional array is formed, the zero values of which correspond to free positions in the original solution matrix. In addition, a special array $E$(1: $nRow$, 1: $nRow$) is created in this block, based on which we can determine the number of free positions in the remaining free rows that will be closed if we select the position ($i$, $j$) to set the queen in solution matrix. Before transferring control to $Block$-5, the following actions are performed:
\\
\newline\indent a) the amount of free positions in all remaining rows is determined
\\
\newline\indent b) the array of free positions, for the considered rows, is ranked in ascending order
\\
\newline\indent c) if all the remaining free rows have free positions (i.e. the minimum value in this ranked list is different from 0), then control is transferred to $Block$-5. If it turns out that in any of the remaining rows there is no free position, then the necessary arrays are restored based on the stored copies, and, depending on the parameters of the task, control is transferred to one of the basic levels.
\\
\newline\indent d) backup copies of all control arrays for this 4th level are formed.
\subsection{Block-5}
The last, final stage, and here, the formation of the search branch is performed more "balanced" and "rational." This is the "last mile", only a small number of free rows remain. But at the same time, this is the most difficult part. All errors that could potentially have been made at the previous stages, together appear here - in the form of a lack of a free position in the row.
The algorithm of this block is executed on the basis of two nested loops, inside which the third loop is executed. A feature of the third cycle is that it can be repeated, without changing the parameters of the two external cycles. This happens if the generated search branch is deadlocked. The number of such repetitions does not exceed the boundary value of $repeatBound$, the optimal value of which was established on the basis of computational experiments.
The outer loop index is associated with a sequential choice of row indices that remained free after calculations at the third base level. This is done on the basis of a previously ranked list of rows by the amount of free positions in the row. The selection starts with a row, with a minimum number of free positions and then, in subsequent steps, in ascending order. Inside this cycle, a second cycle is formed, the index of which iterates over the indices of all free positions in the row. The purpose of the first cycle is only to select the index of one of the free rows at this level. Accordingly, the purpose of the second cycle is only to select one free position within the considered row. These actions occur only at the third basic level. After this choice, the number of installed queens is incremented, and the corresponding cells of all control arrays are changed. Further, control is transferred inside the nested (third) cycle, the activity zone of which is already all the remaining free rows. Inside this cycle, the choice of the row index and the choice of a free position in this row are performed based on the following rules:
\\
\newline\indent a) \textbf{Selection of a free row}. All remaining free rows are considered, and the number of free positions is determined in each row. The row is selected for which the number of free positions is minimal. This minimizes the risks associated with the possibility of excluding the last free positions in some of the remaining rows in which the state is critical in terms of the number of free positions (\textbf{minimum risk rule}). Incidentally, it is with this rule in mind that the index of the first cycle in this fifth block begins with sequential selection of rows with a minimum value of the number of free positions in a row. If at some step it turns out that the two rows have the same minimum number of free positions, then the index of one of the two positions listed first in the ranked list is randomly selected. If the number of rows having the same minimum number of free positions is more than two, then the index of one of the three positions listed first in the ranked list is randomly selected.
\\
\newline\indent b) \textbf{Selection of a free position}. From the list of all free positions in the row, one is selected that causes minimal damage to free positions in all remaining rows. This is done based on the array $E$ formed earlier. By "minimal damage", we mean the choice of such free position in a given row that excludes the least amount of free positions in all remaining rows (the \textbf{minimum damage rule}). If it turns out that two or more free positions in a row have the same minimum values according to the damage criterion, then the index of one of the two positions listed first in the list is randomly selected. Choosing a position that excludes the minimum number of free positions in the remaining rows minimizes the "damage" associated with the queen in a given position. Using both of these rules allows more rational use of resources at each step of forming a search branch. This greatly reduces the risks and increases the likelihood of completion an arbitrary composition to a full solution if the composition in question has a solution. If at some step of the solution it turns out that in one of the remaining rows there are no empty positions, then this search branch is closed. In this case, all control arrays are restored on the basis of backup copies, and if the counter of the number of repetitions does not exceed the boundary value $repeatBound$, then without changing the indices of the first and second external cycles, the work of the third nested cycle is repeated again. This is due to the fact that in cases where the minimum values of the relevant criteria coincided, we made a random selection. Re-forming the search branch on the same conditions of the base level allows more efficient use of the "start resources" provided at this level. The number of repeated starts of the third nested cycle is limited, and if the limit value is exceeded, the operation of this cycle is interrupted. After that, the values of the control arrays are restored, and control is transferred to the cycle of the third base level to move to the next index value. This procedure is repeated cyclically until a complete solution is obtained, or it turns out that we used all the free rows and all the free positions in these rows at this basic level. In this case, depending on the total number of repeated calculations at various base levels, and taking into account the size of the solution matrix and the size of the composition, we return to one of the lower levels for repeated calculations, or a decision is made that the composition can't be completed. In the program, in order to limit the total counting time, it is assumed that the Back Tracking procedure, no matter which of the previous levels is returned, can be performed no more than 1000 times ($totSimBound$=1000). This boundary value was determined on the basis of a large number of computational experiments. If necessary, the value of $ totSimBound $ can be changed.
\section{Analysis of the effectiveness of selection algorithms}
\textbf{The effectiveness of the $randSet$ \& $randSet$ algorithm}. For  analyzing the capabilities of this algorithm, a computational experiment was carried out, which consisted of placing queens in the solution matrix based on the $randSet$ \& $randSet$ model as long as such an opportunity exists. As soon as the search branch reached a dead end, or a complete solution was obtained, the composition size and the solution time were fixed, and the test was repeated again. Computational experiments were carried out for the entire base list of $n$ values. The number of repeated tests for the values $n$ = (30, 40, ..., 90, 100, 200, 300, 500, 800, 1000) was equal to one million, for other values, the number of tests, with increasing $n$, gradually decreased from $10^5$ up to 100. Analysis of the results of computational experiments allows us to do the following conclusions:
\\
\newline\indent a) As a result of the work of only the first cycle of the $randSet$ \& $randSet$ procedure, on average, about 60\% of all queens are correctly placed. For $n$ = 100, the number of queens correctly placed is 60.05\%. With increasing value of $n$, this value gradually decreases, and, for $n$ = $10^7$, is 59.97\%.
\\
\newline\indent b) The histogram of the length of the different compositions has the same appearance, regardless of the size of the decision matrix $n$. Moreover, they all have a characteristic feature - the left side of the distribution (to the modal value) differs from the right side. Figure 2, as an example, presents the corresponding histogram for $n$ = 100.
\begin{figure}[h]
\center\includegraphics[width=1.0\textwidth]{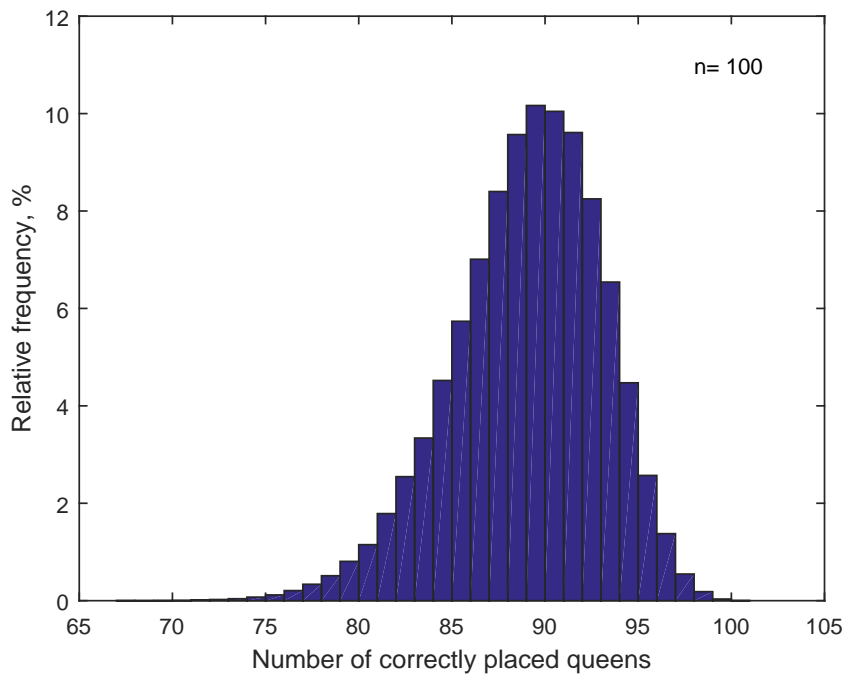} 
\caption{Histogram of various length compositions, formed on the basis of $randSet$ \& $randSet$ model. ($n$ = 100, sample size = $10^6$).}
\label{Fig-2}
\end{figure}
\\
\newline\indent c) The average number of queens ($qMean$) that can be set in the solution matrix based on this algorithm increases with $n$. As can be seen from Figure 3, where a graph of the dependence of the $qMean/n$ ratio on the matrix size $n$ is presented, this ratio increases with an increasing the matrix size. For example, if algorithm for a $100 \times 100$ matrix  allows "endless" to place queens on average 89 rows, then for a $1000\times 1000$ matrix, the number of such rows increases on average to 967.
\begin{figure}[h]
\center\includegraphics[width=0.9\textwidth]{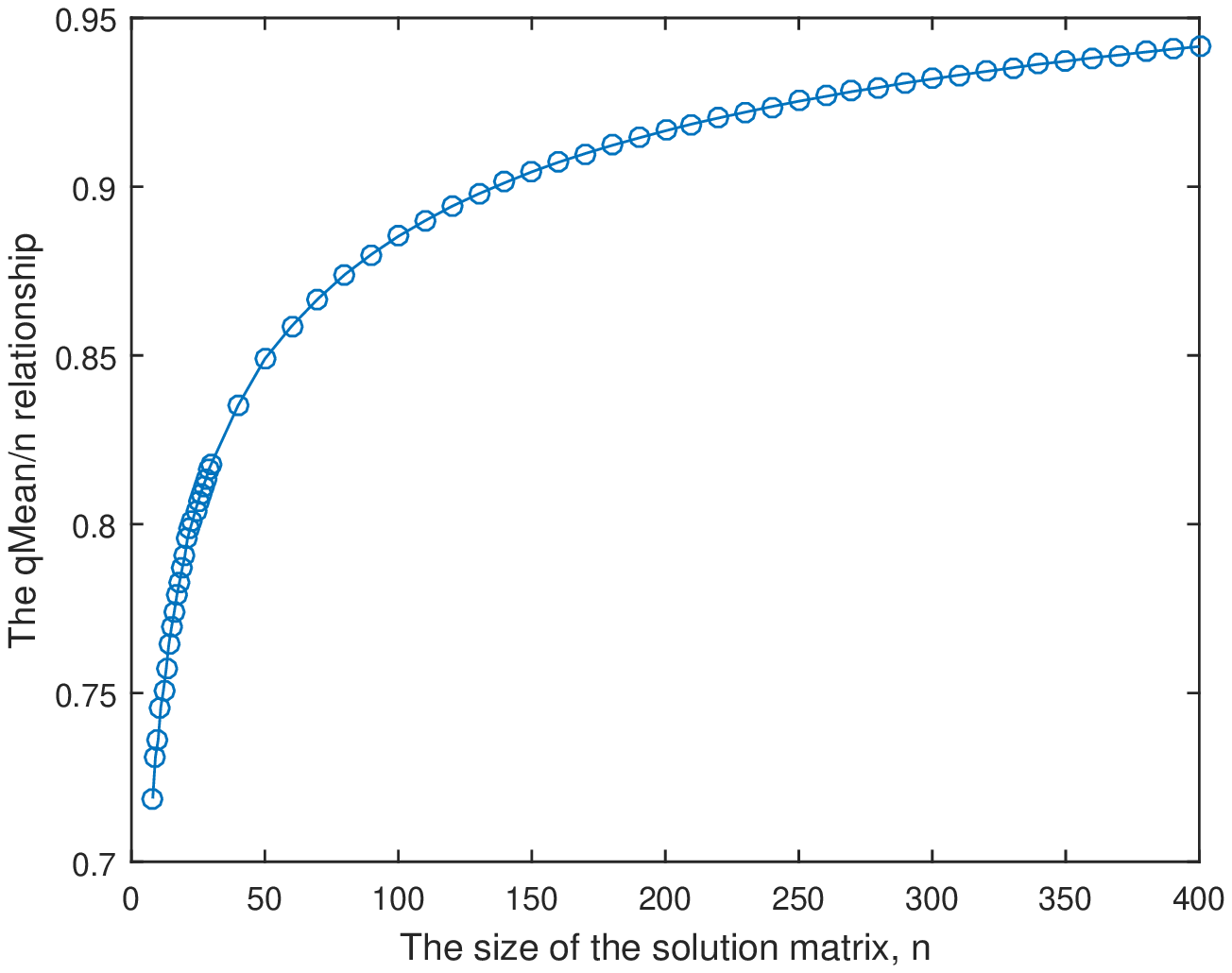} 
\caption{Dependence of the value $qMean/n$ on the $n$ for different sizes of solution matrix (model $rand$ \& $rand$). $qMean$ - average compositions  length.}
\label{Fig-3}
\end{figure}
\\
\newline\indent d) Based on the $randSet$ \& $randSet$ algorithm, a complete solution can be obtained, however the "productivity" of this approach is extremely low. As can be seen from Figure 4, for the value $n$ = 7, the probability of obtaining a complete solution is 0.057. Further, when the value of $n$ is increases, then the probability of such event rapidly decreases, asymptotically approaching to zero. Starting from the value $n$=48, the probability of obtaining a complete solution is of the order of $10^{-6}$. After the threshold value $n$ = 70, for the subsequent values of $n$, not a single complete solution was obtained (for the one million sample size).
\begin{figure}[h]
\center\includegraphics[width=0.9\textwidth]{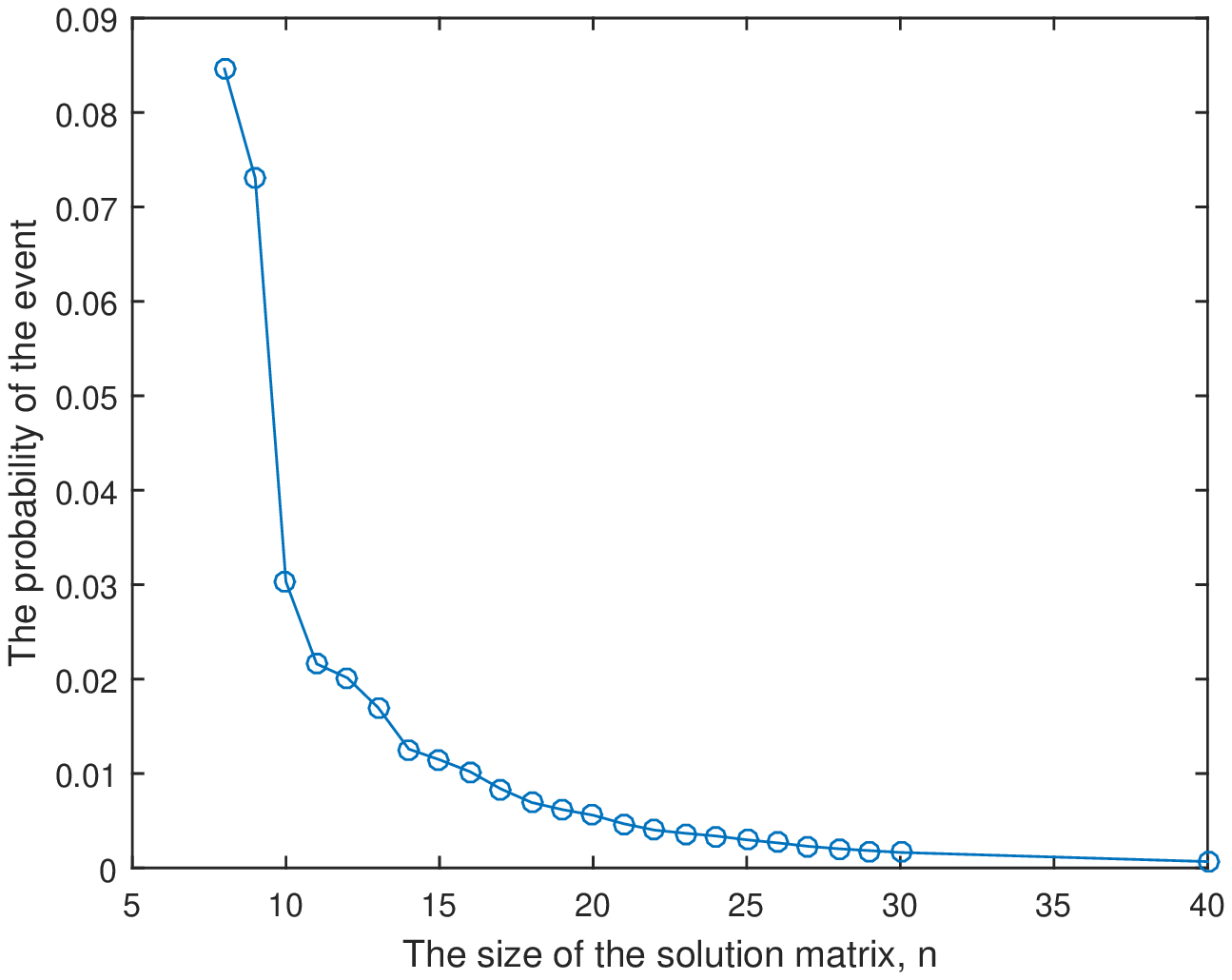} 
\caption{Decreasing the probability of obtaining a complete solution in $randSet$ \& $randSet$ model with increasing n.}
\label{Fig-4}
\end{figure}
\\
\newline\indent e) The $randSet$ \& $randSet$ model generates search branches at a very high speed. For $n$ = 1000, the average time to obtain the composition is 0.001372 seconds. The average length of the compositions is 967. Accordingly, for $n$ = $10^6$, the average time is 1.767318 seconds with an average length of the compositions is 999793.
\\
\newline\indent f) Except for a small interval $n \leq 70$, when the $randSet$ \& $randSet$ model in very rare cases can lead to a solution, in all other cases the solution ends with the formation of a negative composition, which cannot be completed.
\\
\newline\indent Thus, the $randSet$ \& $randSet$ algorithm has an important advantage -  the high speed of forming the search branch, and a significant disadvantage, is that if the composition size exceeds a certain threshold value, then this algorithm leads to form a compositions that cannot be completed. To overcome this drawback, we stop the formation of the search branch when the threshold $baseLevel_2$ is reached.
\\
\newline\textbf{The efficiency of the $rand$ \& $rand$ algorithm}. To determine the capabilities of the $rand$ \& $rand$ algorithm, a fairly detailed computer simulation was carried out for a basic list of $n$ values. As with the $randSet$ \& $randSet$ model, the retest sample size in most cases was one million. For other values of  $n$, the sample size gradually decreased from $10^5$ to 100.
Both algorithms are based on the principle of random selection. Therefore, it should be expected that the conclusions drawn here will be basically identical to the conclusions formulated for the $randSet$ \& $randSet$ model. However, there is a difference between them, and it consists in the following:
\\
\newline\indent a) the $rand$ \& $rand$ model does not work as hard as the $randSet$ \& $randSet$ model. If we talk about some "index of rational use of the provided opportunities", then the $rand$ \& $rand$ model uses resources more rationally at every step. This leads to the fact that, for example, at $n$ = 30, the probability of obtaining a complete solution of 0.00170 in this model is 15 times greater than the similar value of 0.00011 for the $randSet$ \& $randSet$ model. In addition, here, up to the threshold value $n$ = 370, the probability of obtaining at least one complete solution during one million tests remains. After this threshold value, for subsequent values of $n$ with the number of tests equal to one million, not a single complete solution was obtained on the basis of the $rand$ \& $rand$ model.
\\
\newline\indent b) this algorithm is much slower than the $randSet$ \& $randSet$ algorithm. If, for $n$ = 1000, a composition of size 967 is generated, the average time to obtain one composition will be 0.049679 seconds, which is 36 more than the corresponding value of 0.001372 for the $randSet$ \& $randSet$ model.
The reason for the differences between two essentially similar methods of random selection is due to the fact that in the $randSet$ \& $randSet$ model, in order to speed up the calculations, random selection from the remaining list is not carried out at each step. Instead, a pair of indices is sequentially selected from two lists, whose elements were randomly rearranged. Such selection is not random at the full extent, however, it fits well with the logic of the problem and allows us to quickly count.
\\
\newline\indent To visually demonstrate the operation of the $rand$ \& $rand$ algorithm, the following experiment was conducted:
for a chessboard of size $100 \times 100$, after each queen placement step, the number of free positions in each of the remaining free rows is determined. Thus, after each step of solving the problem, we received a list of free rows and a corresponding list of the number of free positions in these rows. Based on these results, a graph was constructed where the indices of the columns of the considered matrix are plotted along the abscissa, and the number of remaining free positions along the ordinates. For comparison, the calculations were also
carried out for the model of sequential selection of positions. By sequential selection is meant the following. The first row is considered, in which the first free position in this row is selected. Then, the second row is considered, in which the first free position in the row is also selected, etc. Figures 5 and 6 show the results that correspond to the models under consideration. For clarity, the graph shows the results only after steps (10, 40, 60). For the model of sequentially selecting positions, the last one is the graph after the 62nd step, since the search branch breaks due to the lack of a free position in the 63rd row. On the other hand, in the $rand$ \& $rand$ model, the last graph is presented after the 70th step of placing the queen, although here, the average number of correctly placed queens reaches 89, which is 26 steps more than in the sequential model. The "strange" type of graphs in the $rand$ \& $rand$ model is due to the fact that the row index is randomly selected from the remaining free rows, and therefore they are randomly scattered throughout the solution matrix.
\begin{figure}[h]
\center\includegraphics[width=0.9\textwidth]{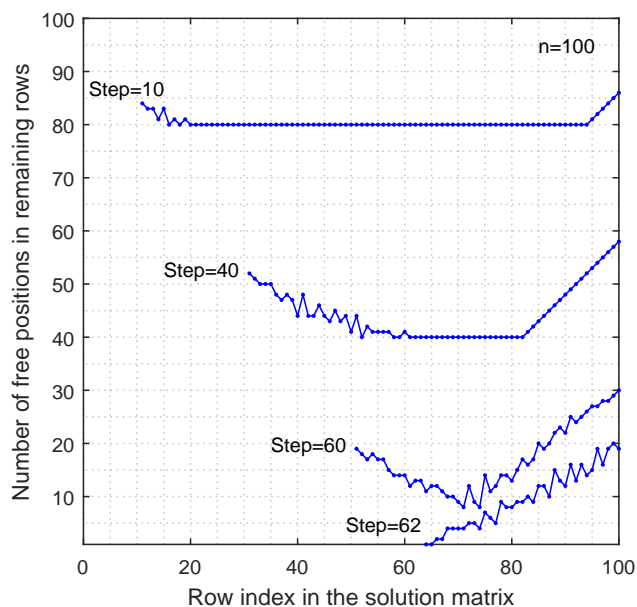} 
\caption{Decreasing the number of free positions in the remaining free rows after placing queens. Sequentially-regular selection of positions.}
\label{Fig-5}
\end{figure}
\begin{figure}[h]
\center\includegraphics[width=0.9\textwidth]{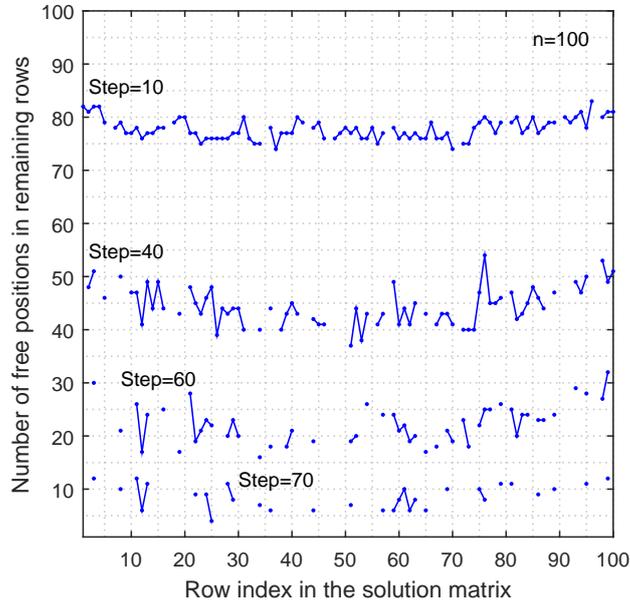} 
\caption{Decreasing the number of free positions in the remaining free rows after placing queens, $rand$ \& $rand$ model.}
\label{Fig-6}
\end{figure}
A comparison of these two figures shows that in a sequential model, the range of variability of the number free positions are higher than in the $rand$ \& $rand$ model. This is due to the fact that with regular selection, diagonal constraints non-uniformly exclude free positions in the remaining rows, which leads to the fact that in some rows the rate of reduction in the number of free positions is higher than in other rows. In contrast, with random selection of the free row index and free column index, the queen's positions are evenly distributed over the "area" of the solution matrix, which reduces the "average" rate of reduction in the number of free positions in the remaining rows. Thus, taking into account the capabilities of the $rand$ \& $rand$ algorithm, we use it in the program to continue the formation of the solution search branch until reach the $baseLevel_3$.
\\
\newline\indent It should be noted that even if the selection algorithms ($randSet$ \& $randSet$, $rand$ \& $rand$) were not so efficient, we would still have to use some other random selection method when developing the algorithm. This is due to the very statement of the n-Queens Completion problem. If we imagine that there is a certain optimal algorithm that solves the problem, then at the input such an algorithm will always receive a random set of row and column indices. Each time it will be a new random set of row and column indices from a huge variety of possibilities. In order to be able to "take in" the algorithm, such a variety of random compositions, the algorithm itself must be built on the basis of random selection. Matching should be like a key to the lock. If we construct the algorithm on this principle, then any consistent composition of $k$ queens will be considered as the initial (starting) position in the solution search cycle. And further, the goal will be only to continue the formation of a branch of the search until a solution for a given composition is found, or it is proved that such a solution does not exist.
\section{An example of using the minimum risk rule ($n$ = 100).}
At the initial stage of the search for a solution, when the number of free positions in the rows is not critical, then the choice of the index of the free row, or the index of the position in this row, is not fatal. However, at the last stage, when the number of free positions in some rows is 1 or 2, we should choose other selection algorithm, since the random selection algorithms $randSet$ \& $randSet$ and $rand$ \& $rand$ will no longer be effective. Let for an arbitrary value of $n$, at some step of solving the problem, in the remaining rows $i_1$, $i_2$, ..., $i_k$, the number of free positions (indicated in brackets) be equal to: $i_1$ (1), $i_2$ (2), $i_3$ (4), $i_4$ (5 ), $i_5$ (3), $i_6$ (4), etc. If we randomly select any row, but not row $i_1$, in which there is only one free position, this can lead to a risky situation where diagonal prohibitions related to the position of the queen in the selected row can lead to the closure of the only free position in row $i_1$ that will lead the solution to a standstill. Of all the rows $i_1$, $i_2$, ..., $i_k$, the most vulnerable and sensitive to the choice of the row index is row $i_1$. In such situations, we should first select the row whose status is the most critical and creates a risk for solving the problem. Therefore, at the last stage of solving the problem, at each step it is necessary to choose the position of the row based on a simple algorithm of minimal risk.
\\
\newline\indent For clarity, let us consider, as an example, for a $100 \times 100$ matrix, the last stage of some solution after the 88th step. Until the task was completed, 12 free rows remained, for each of which the number of free positions was found (the rows are ranked in increasing order of the number of free positions): Step-89 - 25(1), 12(2), 22(2), 82( 2), 88(2), 7(3), 64(3), 3(4), 76(4), 91(4), 4(5), 96(5) - the free row index is indicated, and in brackets - the number of free positions in this row. According to the minimum risk rule, at the 89th step of solving the problem, row 25 is selected and one free position that is in it. As a result of the recount, we have 11 free rows left: Step-90 - 7(2), 12(2), 22(2), 82(2), 88(2), 3(3), 64(3), 76(3), 4(4), 91(4), 96(4). As we can see, the number of free positions in the first five rows is the same and equal to 2. Therefore, the index of one of the first three rows is randomly selected. In this case, the 12th row was selected and the position of the two remaining in this row, which leads to minimal damage. Thus, at the 91st step of forming the solution, we have 10 free rows: Step-91 - 22(1), 3(2), 7(2), 82(2), 88(2), 64(3) 76(3), 91(3), 4(4), 96(4). At this step, row 22 is selected and one free position that is in it. Continuing in a similar way, the following sequence of decisions was formed (Table 1). Indices of selected rows are in bold.
\begin{table}[h]
\footnotesize
\tabcolsep=0.2cm
\renewcommand{\arraystretch}{1.2} 
\begin{tabular}{|r|*{12}{r|}}
\hline
Step & row & row & row & row & row & row & row & row & row & row & row & row \\ \hline
\hline
89 & \textbf{25(1)} & 12(2) & 22(2) & 82( 2) & 88(2) & 7(3) & 64(3) & 3(4) & 76(4) & 91(4) & 4(5) & 96(5)\\ \hline
90 & 7(2) &	\textbf{12(2)} & 22(2) & 82(2) & 88(2) & 3(3) & 64(3) & 76(3) & 4(4) & 91(4) & 96(4) & \\ \hline							
91 & \textbf{22(1)} & 3(2) & 7(2) & 82(2) &	88(2) &	64(3) & 76(3) & 91(3) & 4(4) & 96(4) & & \\ \hline						
92 & \textbf{88(1)} & 3(2) & 7(2) & 82(2) & 91(2) & 64(3) & 76(3) & 4(4) & 96(4) & & & \\ \hline						
93 & \textbf{3(1)} & 7(2) & 76(2) & 82(2) & 91(2) & 4(3) & 64(3) & 96(4) & & & & \\ \hline				
94 & \textbf{76(1)} & 4(2) & 7(2) & 82(2) & 91(2) & 64(3) & 96(4) & & & & & \\ \hline
95 & 4(1) & \textbf{91(1)} & 7(2) & 82(2) & 64(3) & 96(3) & & & & & & \\ \hline
96 & \textbf{4(1)} & 82(1) & 7(2) & 64(3) & 96(3) & & & & & & & \\ \hline
97 & \textbf{7(1)} & 82(1) & 64(2) & 96(3) & & & & & & & & \\ \hline
98 & \textbf{82(1)} & 64(2) & 96(2) & & & & & & & & & \\ \hline
99 & 64(1) & \textbf{96(1)} & & & & & & & & & & \\ \hline
100 & \textbf{64(1)} & & & & & & & & & & & \\ \hline
\end{tabular}
\normalsize
\caption{Demo of using the minimum risk rule ($n$ = 100).}
\label{table_1}
\end{table}
\newline\indent In this particular example, in 11 cases out of 12, there was a situation when in the list of remaining free rows there was at least one row in which only one free position remained. If we did not use the minimum risk rule, we would not be able to get to the end. Since one "wrong move" in choosing the index of a free row, it would most likely lead to the destruction of the only free position that existed in one of the remaining free rows. This is the reason that when using only the $randSet$ \& $randSet$ or $rand$ \& $rand$ algorithm to get the complete solution, in the last stages, the solution goes to a dead end.
\newline\indent It should be noted that the minimum risk algorithm has a simple everyday meaning, and is often used in decision-making. For example, the doctor first of all operates on the patient whose condition is most critical for life, similarly, the farmer, during a severe drought, trying to save the crop, first of all watered those areas that are in the most critical condition \ldots
\begin{table}
\footnotesize
\tabcolsep=0.2cm
\renewcommand{\arraystretch}{1.4} 
\begin{tabular}{|*{8}{r|}}
\hline
$n$ & $m$ & $tMean$ & $t90Mean$ & $tMin$ & $tMax$ & $F-N$ & $tRow$\\ \hline
\hline
10 & 5000& \textbf{0.001010} & 0.000532 & 0.000168 & 0.080673 & 2 & 1.0102 \\ \hline									
20 & $10^5$ & \textbf{0.003589} & 0.001809 & 0.000197 & 0.363096 & 5 & 1.7945 \\ \hline							
30 & $10^5$ & \textbf{0.008025} & 0.003793 & 0.000244 & 0.495716 & 10 & 2.6752 \\ \hline							
40 & $10^5$ & \textbf{0.014323} &	0.009127 & 0.000252 & 0.965817 & 7 & 3.5807 \\ \hline						
50 & $10^5$ & \textbf{0.005357} & 0.003589 & 0.000313 & 0.441711 & 9 & 10.7146 \\ \hline				
60 & $10^5$ & \textbf{0.005991} & 0.004103 & 0.000340 & 0.013738 & 10 & 9.9852 \\ \hline				
70 & $10^5$ & \textbf{0.006533} & 0.004566 & 0.000368 & 0.583897 & 8 & 9.3328 \\ \hline 		
80 & $10^5$ & \textbf{0.006975} & 0.004987 & 0.000394 & 0.635676 & 7 & 8.7187 \\ \hline		
90 & $10^5$ & \textbf{0.006912} & 0.004763 & 0.000393 & 1.012710 & 4 & 7.6840 \\ \hline	
100 & $10^5$ & \textbf{0.007264} & 0.005107 & 0.000419 & 0.692387 & 4 & 7.2641 \\ \hline
300 & $10^5$ & \textbf{0.013518} & 0.009496 & 0.000986 & 3.349766 & 3 & 4.5060 \\ \hline
500 & $10^5$ & \textbf{0.028194} & 0.014554 & 0.001541 & 4.558749 & 2 & 5.6388 \\ \hline
800 & $10^5$ & \textbf{0.049385} & 0.022735 & 0.002367 & 6.192782 & 1 & 6.1731 \\ \hline
1000 & $10^6$ & \textbf{0.062157} & 0.027727 & 0.002943 & 8.015123 & 0 & 6.2156 \\ \hline
3000 & $10^5$ & \textbf{0.177290} & 0.088507 & 0.008537 & 16.713140 & 0 & 5.9097 \\ \hline
5000 & $10^5$ & \textbf{0.159239} & 0.136047 & 0.014224 & 42.181080 & 0 & 3.1849 \\ \hline
$10^4$ & $10^5$ & \textbf{0.321003} & 0.270927 & 0.028594 & 79.321174 & 0 & 3.2100 \\ \hline
$3\ast10^4$ & $10^4$ & \textbf{0.968795} & 0.651618 & 0.084936 & 139.28827 & 0 & 3.22937 \\ \hline
$5\ast10^4$ & 5000 & \textbf{1.147196} & 0.864045 & 0.143005 & 154.38225 & 0 & 2.2944 \\ \hline
$8\ast10^4$ & 4000 & \textbf{2.112079} & 1.215612 & 0.229532 & 204.27321 & 0 & 2.6401 \\ \hline
$10^5$ & 2000 & \textbf{2.253118} & 1.433197 & 0.290566 & 224.34623 & 0 & 2.2531 \\ \hline
$3\ast10^5$ & 2000 & \textbf{4.330649} & 3.181905 & 0.990932 & 340.29584 & 0 & 1.4435 \\ \hline
$5\ast10^5$ & 2000 & \textbf{5.985339} & 4.532205 & 1.488209 & 382.20016 & 0 & 1.1971 \\ \hline
$8\ast10^5$ & 2000 & \textbf{8.297512} & 6.554302 & 2.902425 & 75.87513 & 0 & 1.0372 \\ \hline
$10^6$ & 1000 & \textbf{11.376632} & 7.932194 & 2.954968 & 510.6265 & 0 & 1.1377 \\ \hline
$5\ast10^6$ & 300 & \textbf{33.103386} & 28.057816 & 14.937556 & 155.0890 & 0 & 0.6621 \\ \hline
$10\ast10^6$ & 200 & \textbf{61.444001} & 52.269241 & 31.624475 & 228.3087 & 0 & 0.6144 \\ \hline
$30\ast10^6$ & 50 & \textbf{149.71717} & 136.66441 & 84.556686 & 352.0534 & 0 & 0.4991 \\ \hline
$50\ast10^6$ & 40 & \textbf{253.86220} & 228.93732 & 105.37934 & 558.4629 & 0 & 0.5077 \\ \hline
$80\ast10^6$ & 30 & \textbf{372.29294} & 341.56397 & 250.80182 & 728.4806 & 0 & 0.4654 \\ \hline
$100\ast10^6$ & 20 & \textbf{508.43573} & 474.04890 & 354.80864 & 831.3753 & 0 & 0.5084 \\ \hline
\end{tabular}
\normalsize
\caption{Completion of random compositions for different values of $n$.\ (\footnotesize $n$ – solution matrix size; $m$ - compositions sample size; $tMean$, $tMin$, $tMax$ - average, minimum and maximum completion time, s; $t90Mean$ - the sample average time, from which 10\% of the maximum elements of the ranked series are excluded, s; $F-N$ - the False Negative solutions, number of cases when the positive composition was not completed; $tRow$ = $tMean$ $\ast10^6$ $/$ $n$, the average time(s) increased by $10^6$ times, which is necessary for the queen to be placed on one row).}
\normalsize
\label{table_2}
\end{table}
\section{Algorithm Performance Analysis.}
To evaluate the efficiency of the algorithm for various values of $n$, a rather lengthy computational experiment was performed over the total time. Initially, a fairly fast algorithm was developed for generating arrays of solutions of $n$-Queens Problem for an arbitrary value of $n$. Then, based on this program, large samples of solutions were formed for a basic list of n values. The sizes of the obtained samples of $n$-Queens Problem solutions, respectively, were equal: (10) - 1000, (20, 30, ..., 90, 100, 200, 300, 500, 800, 1000, 3000, 5000, 10000) - 10000, (30000, 50000 , 80000) - 5000, ($10^5$,  $3\ast10^5$) - 3000, ($5\ast10^5$, $8\ast10^5$, $10^6$) - 1000, ($3\ast10^6$) - 300, ($5\ast10^6$) - 200 , ($10\ast10^6$) - 100, ($30\ast10^6$) - 50, ($50\ast10^6$) - 30, ($80\ast10^6$, $100\ast10^6$) - 20. Here, in brackets, a list of values of $n$ is indicated, and through a double dash - the obtained sample size. After that, random compositions of arbitrary size were formed on the basis of each sample. For example, for each of 10000 solutions for $n$ = 1000, one hundred random compositions of arbitrary size were formed. As a result, a sample of one million different compositions was obtained. Since any composition of an arbitrary size, formed on the basis of an existing solution, can be completed at least once, the task was to complete each composition from the generated sample to a complete solution based on the $n$-Queens Completion Problem solution algorithm. Since in the algorithm at each step the correct placement of the queen on the chessboard is checked, here, in principle, there can be no False Positive solutions. However, there may be False Negative solutions - in the event that any composition formed on the basis of the existing solution will not be completed to full solution. Carrying out a computational experiment in such a wide range of $n$ values, we set ourselves the following goals: a) to determine the time complexity of the algorithm; b) determine the probability of the occurrence of False Negative solutions; c) determine the frequency with which the Back Tracking procedure is used in solutions. The results of such a computational experiment are presented in table 1.
\newline\indent The general conclusion that can be drawn based on the obtained results is follows:
\\
\newline\indent a) The algorithm works fast enough. For example, the average calculation time of an arbitrary composition for $1000\times 1000$ chessboard, obtained on the basis of one million computing experiments, is 0.062 seconds. This means that if the composition has a solution, then it will be found immediately after pressing the "Enter" key.
\\ 
\newline\indent b) In each sample, there are approximately 10\% of the compositions that require more time to complete. Such compositions form a long right tail in the histogram. If we exclude these 10\% of the compositions and carry out the calculations for the remaining 90\% of the solutions, then the calculation time ($t90mean$) will be much less. For example, for a $1000\times 1000$ chessboard, the average counting time will be 0.0277 seconds, which is 2.24 times less than the average time obtained from the entire sample.
\\
\newline\indent c) For values $n\le 800$, in the sample of compositions there were those that could not be completed. This is a False Negative solutions. Within the limits specified in the program, allowing the Back Tracking procedure to be performed up to 1000 times, the algorithm failed to complete these compositions. They were mistakenly classified as negative compositions, i.e. those that do not have a solution. The number of such False Negative solutions is insignificant, and their share is mainly less than 0.0001 of sample size. As $n$ increases, the share of False Negative solutions decreases. For all values of $n > 800$, in this series of experiments, there was not a single case of False Negative solutions. However, it is obvious that if the sample size is increased many times, the possibility of the appearance of a False Negative solution is not excluded, although the probability of such an event will be a very small number.
\begin{figure}[h]
\center\includegraphics[width=0.9\textwidth]{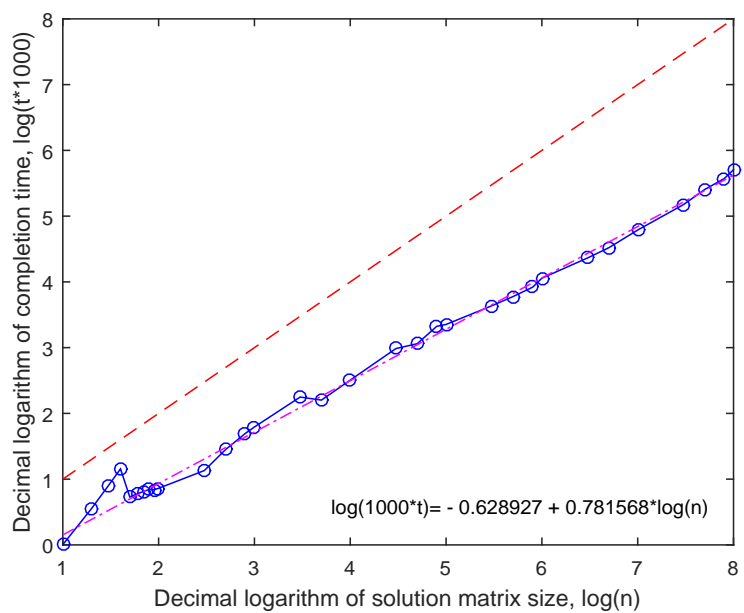} 
\caption{Dependence of the average completion time ($t$) of an arbitrary composition on the size ($n$) of solution matrix.}
\label{Fig-7}
\end{figure}
\newline\textbf{The time complexity of the algorithm}. Figure 7 shows a graph of changes in the average completion time of random compositions for different $n$. The decimal logarithm of $n$ is plotted along the abscissa axis, and the logarithm, increased by 1000 times, of the average counting time, is plotted along the ordinate axis. For clarity, the figure also shows the dotted line of the diagonal of this quadrant. It can be seen that the solution time increases linearly with an increasing $n$. Over the entire range of $n$ values from 50 to $10^8$, the experimental values of the counting time form a straight line, which is described with a fairly high accuracy (R = 0.9998) by the linear regression equation
\\
\newline\indent \textbf{ $log(1000*t) = - 0.628927 + 0.781568*log(n)$}
\\
\newline\indent A slight deviation from the general trend is typical only for the values $n$ = (10,...,49), which is due to the fact that only the fifth block of calculation is used in this range. The algorithm for finding the solution of the fifth block is significantly different from the operation of the algorithms of the first and third blocks.
\par\indent In the obtained dependence, the linear coefficient (0.781568) is less than one, which leads to the fact that with increasing $n$, the regression line and the diagonal of the quadrant diverge.
\begin{figure}[ht]
\center\includegraphics[width=0.9\textwidth]{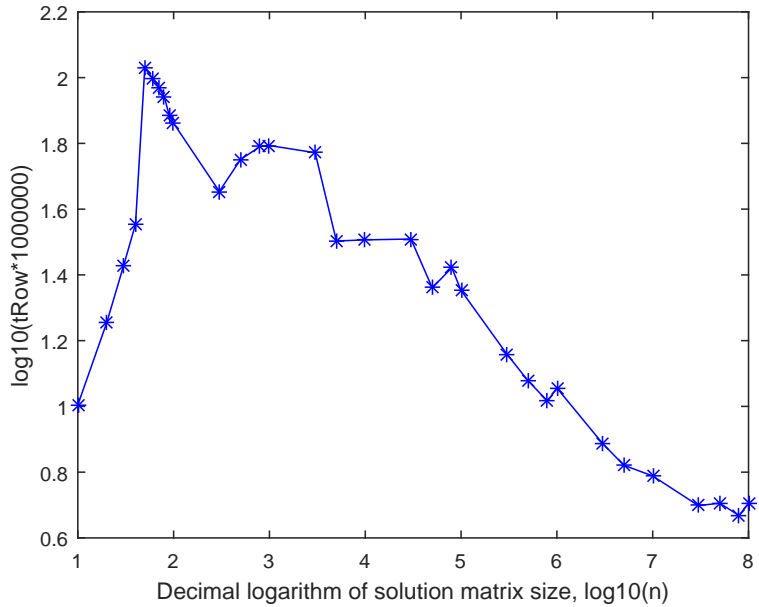} 
\caption{Dependance of the average time to place queen on one row on the size of solution matrix. $tRow$ - average time required to place queen on one row.}
\label{Fig-8}
\end{figure}
\newline\indent In order to clearly explain the reason for this discrepancy instead of the initial time, we consider the average time that is necessary for the location of one queen on one row, i.e. divide the average counting time by $n$. We call such an indicator the reduced time. Obviously, if the reduced time does not change with increasing $n$, then such a solution will be linear ($O(n)$). As can be seen from Figure 8, which shows a plot of the logarithm of the reduced time ($tRow$), increased by $10^6$ times, on the logarithm of the size of the solution matrix, in the range of $n$ from 50 to $10^8$, the reduced time decreases with increasing $n$. If the reduced time for $n$ = 50 is $10.7146 \star 10^{-6}$ seconds, then the corresponding time for $n$ = $10^8$ decreases by 21 times and is $0.5084 \star 10^{-6}$ seconds. Such behavior of the algorithm, at first glance, seems erroneous, since there are no objective reasons why the algorithm will consider it slower for small values of $n$ than for large values. However, there is no error, and this is an objective property of this algorithm. This is due to the fact that this algorithm is a composition of three algorithms that operate at different speeds. Moreover, the number of rows processed by each of these algorithms changes with increasing $n$. It is for this reason that the counting time is increasing in the initial interval of values $n$ = (10, 20, 30, 40), since all calculations in this small area are carried out only on the basis of the fifth block of procedures, which works very efficiently, but not as fast as first block of procedures. Thus, taking into account that the counting time required for placing the queen on one row decreases with increasing size of a chessboard, the time complexity of this algorithm can be called decreasing-linear.
\begin{figure}[h]
\center\includegraphics[width=0.9\textwidth]{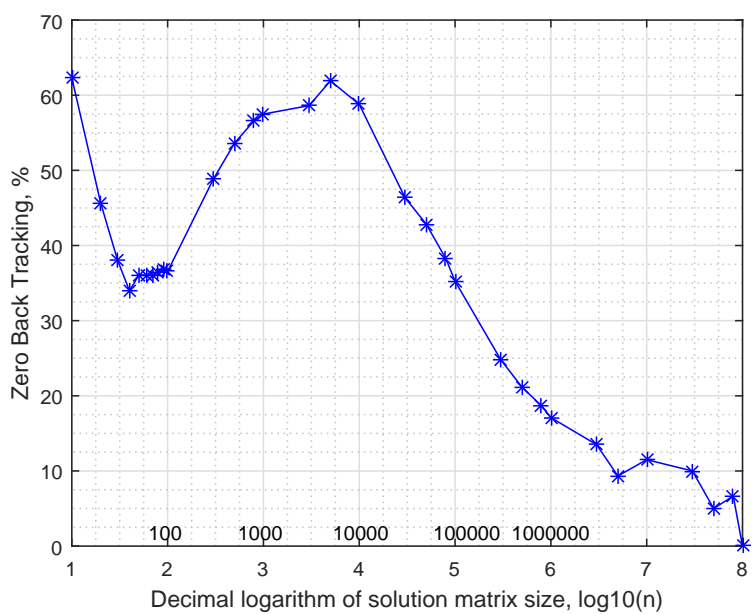} 
\caption{The share of solutions in the sample in which the BT procedure has never been used.}
\label{Fig-9}
\end{figure}
\newline\textbf{The number of cases when the Back Tracking procedure has never been used}. In all cases of a computational experiment, we tracked the number of cases when the Back Tracking(BT) procedure was used in the process of solving each problem. A cumulative summation was made of all cases of using BT, regardless of on what base level was returned to during the search for a solution. This gave us the opportunity to determine for each sample the share of those solutions in which the BT procedure has never been used. Graph in Figure 9 shows how the share of solutions without the use of the BT procedure changes with increasing $n$. It can be seen that in the range of values $n$ = (7,...,100000), the number of solutions in which the BT procedure has never been used exceeds 35\%. Moreover, in the range of values $n$ = (320,...,22500), the number of such cases exceeds 50\%. The most effective results were obtained for a chessboard of size $5000\times 5000$,
where in 61.92\% of cases of completion from 10000 solutions, a "deterministic solution" of a non-deterministic problem was received, because in this cases BT procedure don't been used. In the remaining solutions, in 21.87\% of cases, the BT procedure was used 1 time, in 9.07\% of cases - 2 times, and in 3.77\% of cases - 3 times. Together, this accounts for 96.63\% of all cases. The fact that after the value $n$ = 5000, the number of cases of completion without using the BT procedure is gradually decreasing, is associated with the boundary selecting model of $baseLevel\_2$ and $baseLevel\_3$. We can change these parameters and increase the number of solutions without using the BT procedure. However, this will lead to an increase the counting time, since the share of the fifth block in the algorithm will increase.
\begin{figure}[h]
\center\includegraphics[width=0.9\textwidth]{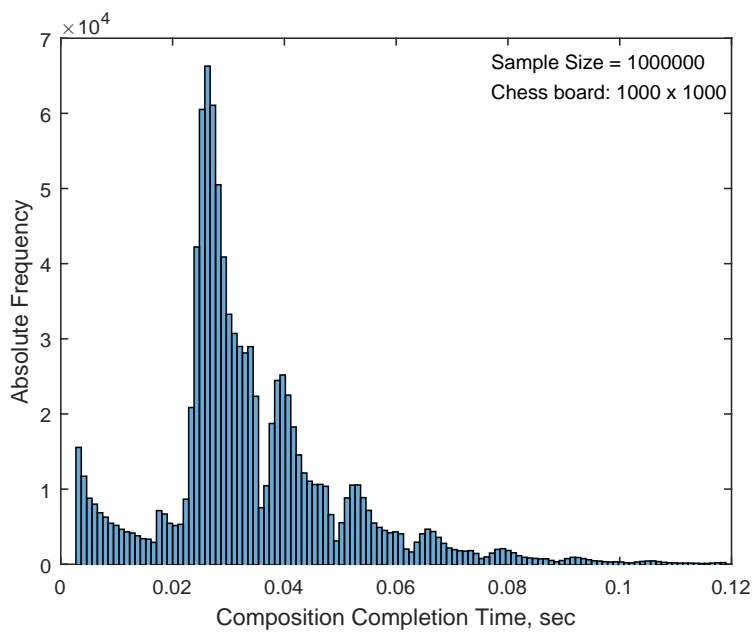} 
\caption{Completion time histogram for compositions with different size. ($n$ = 1000; sample size = $10^6$).}
\label{Fig-10}
\end{figure}
\newline\textbf{A completion time histogram}. In Figure 10, for $n$ = 1000, a completion time histogram for sample of one million solutions is presented. The not quite ordinary view of the distribution histogram (which most likely resembles the night silhouette of high-rise buildings) is not associated with an error in the selection of the length or number of intervals. This is a natural property of this algorithm. To understand why the histogram has such a shape, consider the distribution of completion time for compositions that all have the same size. For this, as an example, we selected from the original sample with size one million, all the compositions whose size is equal to 800. There were 998 such compositions in the sample. Figure 11 shows a completion time histogram for this sub-sample.
It can be seen from the figure that the distribution mainly consists of six separate histograms, with decreasing sizes. The reason why the completion time of 998 compositions, in each of which 800 queens where randomly distributed, "clustered" in 6 groups, is the use of the
Back Tracking procedure. The first histogram in the figure, with the maximum sample size, is those completion solutions where the BT procedure has never been used. This is a group of the fastest solutions. The second histogram, which is significantly smaller in size than the first, is those solutions in which the BT procedure was used only once. Therefore, the solution time in this group is slightly longer than in the first. Accordingly, in the third group, the BT procedure was used twice, in the fourth - three times, etc., i.e. Solutions in which the BT procedure was used repeatedly were performed over a longer time. Such solutions form the long right tail of the desired distribution.
\begin{figure}[h]
\center\includegraphics[width=0.9\textwidth]{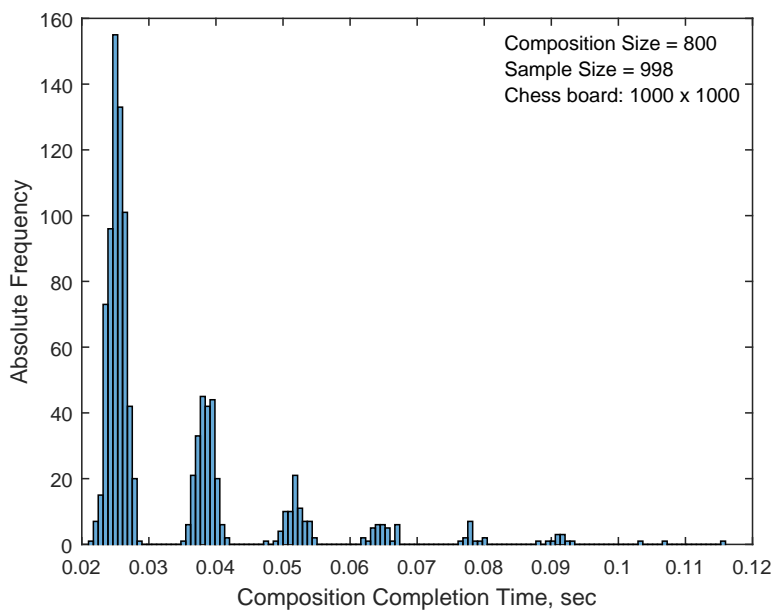} 
\caption{Completion time histogram for compositions with same size. ($n$=1000; compositions size=800; sample size = 998).}
\label{Fig-11}
\end{figure} 
\newline\textbf{False Negative solutions}. If we divide all possible compositions for an arbitrary value of $n$ into positive and negative, then among the positive compositions there are those that this algorithm can classify as negative. This is due to the fact that, within the limits set by the search parameters, the algorithm can't find the right way to complete such compositions. As the experimental results show, the number of such cases does not exceed 0.0001 of sample size, and this error decreases with increasing $n$. In addition, for all values of $n > 800$, there was not a single case of False Negative solution. Even increasing the sample size to one million for a value of $n$ = 1000 did not lead to the appearance of a False Negative solution. The obtained result allows us to formulate the following rule for solving the posed problem: "Any random composition of  $k$ queens that are consistently distributed on an arbitrary chessboard of size $n\times n$ can be completed in linear time, or it will be decided that this composition is negative, and have not solution. The probability of an error in making such a decision does not exceed the value of 0.0001 from the sample size. With the increasing the size of the chessboard, the likelihood of making erroneous decisions decreases."
\section{Conclusions}
1.An algorithm is described which allows solving in linear time the $n$-Queens Completion problem for an arbitrary composition of k queens, consistently distributed on a chessboard of size $n \times n$. Moreover, for any composition of $k$ queens (1 $\le k <n)$, a solution is provided, if any, or a decision is made that this composition can't be completed. The probability of an error in making such a decision does not exceed 0.0001, and this value decreases with increasing size of a chessboard.
\\
\newline 2. The operation of this algorithm is based on the use of two important rules:
\\
\newline\indent a) At the solution final stage, from all the remaining free rows, one is selected for which the number of free positions is minimal (\textbf{ minimum risk rule}). This minimizes the risks associated with the possibility of excluding the last free positions in some of the remaining rows.
\\
\newline\indent b) From all free positions in the row, the position is selected that causes minimal damage to the free positions in the remaining free rows (\textbf{minimum damage rule}). By "minimal damage" is meant the selection of such a position in a row that excludes the least amount of free positions in all remaining free rows.
\\
\newline 3. It is established that, the average time required for the queen to be placed on one row decreases with increasing value of $n$. The average time required for the queen to be placed on one row in the case when $n$ = $10^8$ is 21 times less than the corresponding time for the case $n$ = 50.
\\
\newline 4. It was found that in the range of $n$=(7,...,100000), the number of solutions in which the Back Tracking procedure has never been used exceeds 35\%. Moreover, in the range of values $n$ = (320,...,22500), the number of such cases exceeds 50\%.
\\
\newline 5. A model for organizing the Back Tracking procedure is proposed based on the division of the solution matrix into basic levels. By a level is meant a certain solution step with a given number of correctly placed queens. Regression formulas are given for calculating the values of the second and third basic levels depending on $n$.
\\
\newline 6. The results of a comparative analysis of two random selection methods, which are called $randSet$ \& $randSet$ and $rand$ \& $rand$, are presented. The $randSet$ \& $randSet$ algorithm has been found to be fast, but its use should be limited when reaching the second base level. After that, the $rand$ \& $rand$ algorithm is used, which is not so fast, but more efficiently places queens on the chessboard.
\\
\newline 7. An effective algorithm for verifying the correctness of the $n$-Queens Problem solution is given. This program is also designed to verify the correctness of a composition of arbitrary size. The program works fast enough. For example, the time required to validate a solution consisting of 5 million positions is 0.85 seconds.
\section{Remarks}
1. The algorithm is independent of the programming language. Matlab, in this case, is selected as a environment for modeling, due to ease of use, speed and variety of features. For this, I would like to thank the MathWorks development team. The program uses only the simplest language operators. Therefore, translating script into another programming language will not be a difficult task.
\\
\newline 2. Research to find a solution to this problem continued for a long time. And although, during this period, several tens of millions of solutions were received, the situation is not ruled out when the program, under some conditions, can not cope with the task. I would be glad if you let me know.\
\\
\newline 3. As indicated at the beginning of the article, investigations were conducted in the range of $n$ values, from 7 to 100 million. However, the program was tested in a wider range of $n$ values, up to one billion. True, in the latter case, I had to slightly adapt the program, given the large size of the arrays. Therefore, if the size of the RAM allows, then it is possible to carry out calculations for large values of $n$.\
\\
\newline 4. The values of the baseline indicators, as well as the boundary values of the number of repetitions at various levels, were optimized to solve the problem within the entire range of studies. They can be changed within shorter intervals and achieve a reduction calculation time. It is important not to increase the share of False Negative solutions.\
\\
\newline 5. In this article, I used the Enter keypress time as a measure of time to evaluate how fast the algorithm works. If the result appears immediately after pressing the key, then at the level of user perception, it seems that the program works "very" fast. No matter how fast the algorithm works, the result will appear on the screen no earlier than the time the key is pressed. Therefore, it seemed to me that such a conditional measure of time can serve as a threshold level for not strictly comparing the speed of various algorithms.
\\
\newline 6. During the development an algorithm for solving the n-Queens Completion Problem, another problem was considered. Its essence was to minimize the number of cases using the BT procedure. In all things being equal, the less the BT procedure is used, then the more efficient the algorithm will be. To achieve this, it is necessary to reduce errors, both when choosing the index of a free row,  and in choosing a free position in this row.  This is especially important at the last stage of solving the problem, where any "wrong move" will be fatal for the formed branch of the solution. The fact that in the interval of values n = (320,...,22500) it was possible in more than 50\% of cases to solve the problem without using the BT procedure, suggests that we are quite close to this goal. If we go further along this path, then one of the directions for further algorithm improvement may be search a rule that would give preference to one of the rows, if they have the same minimum number of free positions. However, it may turn out that the time required to calculate a certain functional for all free rows will be higher than necessary for fulfillment the BT procedure and recounting.
\\
\newline 7. \textbf{Philosophical}\ldots In the course of the study, a large number of publications were considered related to the solution of non-deterministic problems. In most cases, these were tasks in which it was necessary to make choice in a large state space under the conditions of given constraints. Comparing them, it was interesting to know how far one can advance in solving such problems using the standard mathematical approach. I got the impression that only on the basis of definitions, statement of lemmas and proof of theorems, it is impossible to solve such problems. It seems to me that to solve such problems it is necessary to use methods of algorithmic mathematics using Computational Simulation. To demonstrate the validity of this conclusion, as a simple example, I prepared for a chessboard of size $10^9$ x $10^9$, two compositions of the same size, consisting of 999 999 482 queens. This data are prepared as described at the beginning of the article and presented as two files in .mat format. They can be downloaded from this link \href{https://drive.google.com/drive/folders/1p5VJO4fAEs8vNp_EonFIitd1o6EJqzQt?usp=sharing}{Two test files}. Files are quite "heavy", the size of each of them is about 3.97 Gb. In 999 997 976 rows (99.9998\%) the positions of the queens in both compositions coincide, and only in arbitrary 1506 lines the positions of the queens differ. To complete this compositions to a full solution, we need to correctly place the queens in the remaining 518 free rows. The number of possible ways to arrange 518 queens in the remaining free rows (taking into account only the number of ways to select a free position in the row) is approximately $10^{1466}$. The difference between these two compositions is only that one of them is positive and can be completed, and the other composition is negative - it cannot be completed. Question: "Is it possible, on the basis of a rigorous mathematical approach (that is, without carrying out algorithmic computational operations), to determine which of these two compositions is positive?" If this is impossible to solve, then we can assume that the proposition made proved by contradiction.
\\
\newline I want to note that no matter what the approach to the strictly mathematical solution of this problem, one needs to determine the status of $518\times10^9$ cells in the remaining free rows. To do this, it is necessary to consider each position of previously established queens, and there are almost one billion of them, to establish the restrictions that each established queen imposes on free positions in the remaining 518 rows. I did not find a "fulcrum" that would allow us to do this work only on the basis of a strictly mathematical approach, without algorithmic calculations.\
\\
\par I have given here a minimal example consisting of only two compositions. If necessary, the number of such compositions can be increased.
It should be noted that, on the basis of the proposed linear algorithm, slightly adapted for large compositions, the tasks of which of the two compositions can be completed is performed on desktop-13, in about 4.5 minutes (excluding data input load time).\
\section{Addition}
The action of professors who recommend difficult tasks for development and research to capable students is worthy of respect. This requires considerable effort, but overcoming difficulties, the researcher looks at other complex tasks differently. I thought that it would be useful to expand the options for staging the $n$-Queens Problem for such purposes. Looking at the same task from different positions, we can see different things. Below are some of them.
\\
\newline 1. Consider the problem of arranging $n$ queens on a rectangular "chess" board of size $n \times m$. Denote $k = m - n$. Let some solution be obtained, and each of the $n$ rows contains one queen. We exclude the positions where the queens are located from further consideration. Now in each row there is $m-1$ free position. Within the remaining free positions, we again find one solution. As before, we exclude from further consideration the positions where the queens of the second solution are located. Now in each row there are $m-2$ free positions. Obviously, the first and second solutions do not intersect in their positions in any row - they are orthogonal. It is required to determine the maximum number of mutually orthogonal solutions for various values of $k$. If $n$ mutually orthogonal solutions are found for the value $k$ = 0, then a Queens Latin Square will be constructed.
\\
\newline\textbf{Comment}. In the publication Grigoryan \cite{Grigoryan-2018} showed that for any n-Queens Problem solution, there is a complementary solution that does not intersect with it. This means that for an arbitrary value of $n$, the set of all $n$-Queens Problem solutions is divided into two equal-sized subsets. Any solution from the one subset is a complementary solution to the corresponding solution from the other. The rule is quite simple, if $Q1$ (i) is a solution from the first subset, then the corresponding complementary solution $Q2(i)$ from the second subset is determined by the formula $Q2(i) = n + 1 - Q1 (i)$, where $i$ = (1,..., $n$). It is this rule that explains the fact that the number of all solutions of the $n$-Queens Problem, for any arbitrary value of $n$, is always an even number.
\\
\newline 2. In the initial formulation of the $n$-Queens Problem, after the queen is placed in position $(i,j)$, the following actions are performed:\
\\
\newline\indent a) exclude all cells of row i and column j
\\
\newline\indent b) all cells that are located on the line of the left and right diagonals passing through the cell $(i,j)$ are excluded.
\\
\newline We change condition b) in the statement of the problem. Instead of eliminating cells, we will use cell switching. If the cell located on the line of the left or right diagonals is free, then we will close it; if the cell is closed, then we will open it. This makes it easier to find a solution. However, instead of a square matrix $n\times n$, we consider a rectangular matrix of size $n\times(n - k)$. It is required, for a given value of $n$, to find the maximum value of $k$ at which at least $h$ orthogonal solutions can be obtained(h=3,4,5). How will the value of $k$ change with increasing value of $n$?
\\
\newline 3. Change some conditions in the initial formulation of the $n$-Queens Problem. When the queen is placed in position $(i,j)$ on a chessboard of size $n\times n$:
\\
\newline a) Exclude all cells in row $i$.
\\
\newline b) If index $j$ is an even number, then
\\
\newline\indent b1) exclude cells in even rows of column $j$,
\\
\newline\indent b2) exclude cells in even lines intersecting with the left and right diagonals passing through the cell $(i,j)$.
\\
\newline c) If index $j$ is an odd number, then points b1) and b2) are satisfied for cells located on odd rows.
\\
\newline\indent 3.1 It is known \cite{Sloane-2016}, that the list of values of all solutions of the nQueens Problem, for $n$ = (8, 9, 10, 11, 12, 13, 14, 15, 16), respectively, is (92, 352, 724, 2680, 14200, 73712, 365596 , 2279184, 14772512). How will the number of all solutions change if, in the statement of the problem, the standard condition for diagonal exceptions is changed to paragraph b) or paragraph c)?
\\
\newline\indent 3.2 It is known \cite{Grigoryan-2018}, that if we determine the frequency of participation of various cells of the solution matrix in the formation of a list of all solutions, we can find that there are harmonious relationships between all cells in the form of vertical and horizontal symmetries of the corresponding frequencies. This means that, if we assume that $k<n$/2, then the frequency of the cells of the $k$-th row will be identical to the frequencies of the cells of the row $n-k$ + 1. Similarly, the frequency of the cells of the $k$-th column will be identical to the frequencies of the cells of the column $n-k$ + 1. Question: How will these harmonious relations change in the context of the task?
\\
\newline 4. All cells of a chessboard are divided into two classes by their color. It is believed that one color is white, the other is black. Consider two chessboards and place one of them on the other so that the edges coincide completely. As a result, we get a "sandwich" of two chessboards in which the arrangement of white and black cells coincide. The task is to find h mutually orthogonal solutions that are located on the first or second board ($h$ = 3,4,5,...), observing the following conditions:
\\
\newline a) If in one of the boards the queen is located on a black cell with indices $(i,j)$, then:
\\
\newline\indent a1) on this board:
\\
\newline\indent\indent - all black cells that are found on row $i$ and column $j$ are excluded,
\\
\newline\indent\indent - all black cells that are located along the left and right diagonals passing through the cell $(i,j)$ are excluded.
\\
\newline\indent a2) on other board:
\\
\newline\indent\indent - all white cells that are found on row $i$ and column $j$ are excluded,
\\
\newline\indent\indent - all white cells that are located along the left and right diagonals passing through the cell $(i,j)$ are excluded.
\\
\newline b) If on one of the boards the queen is located on a white cell with indices $(i,j)$, then all the actions of items a1) and a2) are performed for cells of the opposite color.
\\
\newline 5. Imagine that in a solution matrix of size $n\times n$, rows can slide right or left relative to each other, in increments of k cells. Moreover, if the previous row was shifted, for example, to the left, then the next row should be shifted to the right, i.e. each next row is shifted in the opposite direction to the previous row. As a result of this construction, we obtain a rectangular matrix of size $n\times (n + k)$, where in each row $k$ cells from the beginning of the row or from the end will be excluded from consideration. The problem is to find the maximum value of $k$ for an arbitrary value of $n$ for which there is at least $h$ solution n-Queens Problem ($h$=1,2,3,4,5).
\\
Consider a variant of the problem in which the offset of one line with respect to another is a random number ranging from $k1$ to $k2$.
\\
\newline 6. \textbf{The one-dimensional formulation of $n$-Queens Problem}. Suppose that on some line $n$ segments are numbered from 1 to $n$. Divide each segment into $n$ cells, and within each segment, number cells from 1 to $n$. We call such cells open. It is required to select one open cell in each segment, taking into account the following constraints:
\\
\newline a) We can select the open cell with number $j$ from the $i$-th segment, if:
\\
\newline\indent $D1(r)$ = 0;
\\
\newline\indent $D2(t)$ = 0;
\\
\newline where $r$ = $n + j - i$; $t$ = $j + i$, $D1$ and $D2$ - 1-dimensional control arrays consisting of $2n$ cells that were previously zeroed.
\\
\newline b) After a such selection, will be closed the segment $i$ and cells with number $j$ in all remaining free segments. Also, it is necessary to close the corresponding cells in the control arrays:
\\
\newline\indent $D1(r)$ = 1;
\\
\newline\indent $D2(t)$ = 1;
\\
\newline\indent In this formulation, the task is completely identical to the original one. Of interest are the formulations of this problem with other constraint conditions. For example, if instead of the formulas: $r$ = $n + j - i$ and $t$ = $j + i$, other relations will be considered that functionally connect the indices $r$ and $t$ with the indices $(i,j)$ of the solution matrix.
\\
\newline 7. \textbf{The formulation of the task on the basis of an urn with balls} (identical to the previous task). Suppose there are $n$ urns numbered from 1 to $n$, and in each urn there are $n$ balls, also numbered from 1 to $n$. It is required to select one ball from each urn, given the following restrictions:
\\
\newline a) We can select a ball with number $j$ from the $i$-th urn, if:
\\
\newline\indent $D1(r)$ = 0;
\\
\newline\indent $D2(t)$ = 0;
\\
\newline where $r$ = $n + j - i$; $t$ = $j + i$, $D1$ and $D2$ - 1-dimensional control arrays consisting of $2n$ cells that were previously zeroed.
\\
\newline b) After a such selection, will be closed the urn $i$ and balls with number $j$ in all remaining free urns. Also, it is necessary to close the corresponding cells in the control arrays:
\\
\newline\indent $D1(r)$ = 1;
\\
\newline\indent $D2(t)$ = 1;
\\
\newline In this formulation, the task is completely identical to the original one. As in the previous case, the formulation of this problem with other constraint conditions is of interest.
\\
\newline 8. \textbf{The game}. Consider a chessboard of size $n\times n$. We will return the color to the queens, let some queens have white color, others black. We also return the alternating white and black color to the cells of the chessboard, based on the fact that in solution matrix the cell with the index (1,1) should be white. All cells at the beginning of the game are considered free. One player places white queens, the other blacks. The player places the queen on an arbitrary free cell with indices $(i,j)$. Each player can make a move on both white and black cell. White queens make the first move. Let it be a white cell. As a result of this choice, the following cells are closed:\
\\
\newline\indent a) all white cells of row $i$;\
\\
\newline\indent b) all white cells of column $j$;\
\\ 
\newline\indent c) all white cells that lie on the left and right diagonals passing through the cell $(i,j)$.\
\\ 
\newline If the cell $(i,j)$ turns out to be black, then all items a), b), c) are satisfied, and accordingly, all the cells in black are closed. Further, Black performs the move, placing the queen on any of the remaining free cells. After that, in a similar way, the cells close, as described above. The time to think about the next move is fixed, and is selected by agreement of the parties. If during the specified time, one of the players does not complete his move, then the game is transferred to the other. The game ends if both players do not have a decision to complete the move. The one who can place more queens on the board wins.\
\\
\newline 9. \textbf{The stability of random selection}. Consider the $randSet$ \& $randSet$ model. As a result of comparing $n$ random pairs of row and column indices, at the first stage of the cycle, the queen can be established on average on $k \times n$ rows. The value of $k$ can be considered as a constant value equal to 0.6. Its value varies from 0.605701 at $n$ = 10, and to 0.599777, at $n$ = $10^6$, and, with an increase in $n$, the dispersion around the average value decreases. What is the reason for such "constancy"? Why, with a random selection of the row index and the position index, on the basis of two lists of randomly rearranged numbers, it is possible to consistently place 60\% of queens?\
\\
\newline 10. Let the size of the chessboard be $n\times n$. Based on the $randSet$ \& $randSet$ procedure, we place the queens on the chessboard until the search branch reaches the dead end. If, for a given value of $n$, repeat this procedure many times, and construct a histogram of the distribution of values of $k$, it turns out that the change of events frequency to the value of the distribution mode differs from the change of events frequency after this value. If, based on the modal value, the histogram is divided into two parts, then the left part will not coincide with the right part. This pattern is characteristic for any value of $n$. Why, after the transition of the length of the composition through the modal value, does the frequency of events take a different form? By event, we mean the formation of a composition until a deadlock state is reached.\
\\
%
%

\bibliography{Linear_Algorithm_for_Solving_n_Queens_Completion_Problem_02_dec_2019.bbl}
\end{document}